\documentclass[conference,9pt]{IEEEtran}

\usepackage{array}
\usepackage{algorithm}
\usepackage{amsmath}
\usepackage{amsfonts}
\usepackage{booktabs}
\usepackage{float}
\usepackage{graphicx}
\usepackage{listings}
\usepackage{mathrsfs}
\usepackage{multirow}
\usepackage{multicol}
\usepackage{mwe}
\usepackage{pifont}
\usepackage{siunitx}
\usepackage{soul}
\usepackage[countmax]{subfloat}
\usepackage{subcaption}
\usepackage{xcolor}
\usepackage{booktabs}
\usepackage{array}

\usepackage{url}

\IEEEoverridecommandlockouts

\usepackage{cite}
\usepackage{amsmath,amssymb,amsfonts}
\usepackage{algorithmic}
\usepackage{graphicx}
\usepackage{textcomp}
\usepackage{xcolor}

\def\BibTeX{{\rm B\kern-.05em{\sc i\kern-.025em b}\kern-.08em
    T\kern-.1667em\lower.7ex\hbox{E}\kern-.125emX}}

\begin{document}

\title{CROP: \underline{C}ircuit \underline{R}etrieval and \underline{O}ptimization with \\ \underline{P}arameter Guidance using LLMs}

\author{
Jingyu Pan$^{1,\dag}$\thanks{$^\dag$ Jingyu Pan and Isaac Jacobson contributed equally to this work.}, 
Isaac Jacobson$^{1,\dag}$, 
Zheng Zhao$^{2}$, 
Tung-Chieh Chen$^{2}$,
Guanglei Zhou$^{1}$,
Chen-Chia Chang$^{1}$, \\
Vineet Rashingkar$^{2}$, 
Yiran Chen$^{1}$\\
$^{1}$\textit{Electrical and Computer Engineering, Duke University, Durham, USA}\\
$^{2}$\textit{Synopsys Inc., Sunnyvale, USA}\\
$^{1}$\{jingyu.pan, isaac.jacobson, guanglei.zhou, chenchia.chang, yiran.chen\}@duke.edu\\
$^{2}$\{zheng.zhao, donnie.chen, vineet.rashingkar\}@synopsys.com
}

\maketitle

\begin{abstract}

Modern very large-scale integration (VLSI) design requires the implementation of integrated circuits using electronic design automation (EDA) tools.
Due to the complexity of EDA algorithms, the vast parameter space poses a huge challenge to chip design optimization, as the combination of even moderate numbers of parameters creates an enormous solution space to explore.
Manual parameter selection remains industrial practice despite being excessively laborious and limited by expert experience.
To address this issue, we present CROP, the first large language model (LLM)-powered automatic VLSI design flow tuning framework.
Our approach includes:
(1) a scalable methodology for transforming RTL source code into dense vector representations,
(2) an embedding-based retrieval system for matching designs with semantically similar circuits, and
(3) a retrieval-augmented generation (RAG)-enhanced LLM-guided parameter search system that constrains the search process with prior knowledge from similar designs.
Experiment results demonstrate CROP's ability to achieve superior quality-of-results (QoR) with fewer iterations than existing approaches on industrial designs, including a 9.9\% reduction in power consumption.

\end{abstract}


\section{Introduction}

Modern industrial VLSI design flows exhibit exceptional complexity.
A typical EDA flow comprises multiple sequential stages, with hundreds or even thousands of functions that depend on parameters which need to be configured.
To optimize for an entire VLSI design flow, synthesis-related options in front-end tools like Synopsys Design Compiler or placement, clock tree synthesis, and routing options in back-end tools such as IC Compiler II must be configured appropriately to optimize for a specific QoR target~\cite{geng2022ptpt}.
The explorable parameter space grows exponentially with the number of configuration functions, making it practically impossible to exhaustively search for the best parameter combination given a chip design.
Besides, the parameter configurations dramatically influence the QoR of VLSI designs across all metrics.
For example, experiments showed that minor changes in synthesis parameters can lead to substantial variations in performance, power and area (PPA)\cite{xie2020fist,geng2022ppatuner,zheng2023boosting}.

Consequently, the automation of parameter tuning offers significant advantages for VLSI development.
However, research in this field remains limited due to challenges in acquiring comprehensive design flow datasets necessary for implementing parameter tuning experiments.
\cite{ziegler2016synparam} developed an automatic parameter tuning approach using genetic algorithms that tests various parameter configurations.
\cite{ma2019cad} leverage surrogate Gaussian process models to efficiently explore the design space by capturing correlations among training samples.
\cite{akiba2019optuna} proposed Optuna, a general purpose hyperparameter optimization framework that allows dynamic construction of parameter search spaces during runtime.
Recommender~\cite{kwon2019learning} trained a latent factor model with archived design data to recommend the best parameter configurations of a new design.
\cite{agnesina2020vlsi} proposed a reinforcement learning-based method to suggest the best parameter configuration of a new design in one shot.
FIST~\cite{xie2020fist} automatically learns parameter importance from legacy designs and uses this information for feature-importance sampling and tree-based parameter tuning in new designs.
However, these methods typically require a significant number of sample runs to identify optimal settings, as they lack any understanding of the circuit being optimized.
The fundamental limitation lies in their lack of ability to leverage semantic understanding of the circuit being optimized, which leads to extra effort spent exploring regions of the parameter space that may be irrelevant for the target design.

\begin{table}[t]
\centering

\caption{Comparison of parameter tuning methods. CROP achieved circuit-aware and automatic parameter optimization with high efficiency.}
\begin{tabular}{cccc}
\toprule
\textbf{Method} & \textbf{Circuit-Awareness} & \textbf{Automation} & \textbf{Efficiency} \\
\midrule
\multicolumn{1}{l}{\textbf{Human Effort}} & Yes & No & Low \\ 
\multicolumn{1}{l}{\textbf{Classical Algorithms}} & \multirow{2}{*}{No} & \multirow{2}{*}{Yes} & \multirow{2}{*}{Low} \\
\multicolumn{1}{l}{\cite{liang2021flow,hsiao2024fasttuner,xie2020fist, xu2024rank}} & & & \\ 
\multicolumn{1}{l}{\textbf{CROP}} & Yes & Yes & High \\ 
\bottomrule
\end{tabular}

\label{tab:introduction}
\end{table}

To address this challenge, a promising research direction leverages circuit characteristics to enhance parameter tuning.
This circuit-aware approach incorporates prior knowledge of design structure and functionality to more efficiently constrain the optimization process.
VLSI design flows implementing such circuit-aware parameter tuning have demonstrated significant improvements in search efficiency.
\cite{papamichael2015nautilus} used knowledge provided from circuit designers to guide the genetic algorithm and improve search efficiency.
\cite{zheng2023boosting} used random embedding to convert the parameters to a lower-dimensional latent space, which can be more efficiently explored based on design similarity.
\cite{liang2021flow} exploited archival design data to improve the warm-up efficiency of the search engine.
Despite these advances, significant challenges remain in effectively transferring knowledge in a way that better captures the correlation between pre-existing archival design data and the design currently being optimized.
This is due to insufficient research on the effective encoding of the semantic representations of circuit designs.
An approach that can understand circuit semantics and leverage knowledge from similar designs could dramatically reduce the number of iterations required to find optimal parameters.

Building on recent advances in LLMs and their demonstrated effectiveness in EDA and chip design tasks~\cite{pan2025survey,yao2024rtlrewriter,chen2024dawn}, we investigate the potential of LLMs to enable fully-automated circuit-aware parameter search engines.
Unlike traditional block-box optimization tools, LLMs can be conditioned using rich descriptive prompts that incorporate the QoR target, the functionality of each tool option, and even high-level descriptions of the circuit under optimization.
This allows LLMs to emulate the intuition and practice of experienced engineers, making context-aware adjustments to EDA tool configurations, design constraints, and metric-oriented strategies.
Furthermore, RAG emerges as a systematic approach that augments general-purpose LLMs with a domain-specific knowledge database, enabling more informed parameter optimization decisions based on historical design data and established optimization patterns.
The integration of RAG with LLMs creates a parameter tuning framework that leverages contextual reasoning with circuit-specific prior knowledge, enabling parameter optimization that adapts to unique design characteristics across diverse technology nodes.

In this work we present \textbf{CROP}, the first LLM-powered automatic VLSI design flow tuning framework.
Table~\ref{tab:introduction} summarizes CROP's advantage compared with human effort and classical parameter tuning algorithms.
The main contributions of this paper are listed as follows:
\begin{itemize}
    \item A novel sub-framework for scalable summarization of a design from its RTL source code to a dense vector representation.
    \item An embedding-based design retrieval system to match designs with those that are semantically and structurally similar.
    \item A RAG-enhanced LLM guided automated search system that combines prior knowledge encoded in the LLM and that extracted from the design retrieval system.
    \item CROP finds better QoR than the baseline methods which even failed to explore the parameter space with better QoR values given a limited search budget. Specifically, CROP achieved a \textbf{9.9\%} reduction of the power consumption on an industrial processor design compared to the best of the baselines.
\end{itemize}

The remainder of this paper is organized as follows.
Section~\ref{background} provides relevant preliminaries.
Section~\ref{problem_formulation} contains our problem formulation.
Section~\ref{method} illustrates our methodology.
Section~\ref{experiments} presents experimental results.
Section~\ref{discussion} contains additional discussions.
And Section~\ref{conclusion} presents our conclusion.

\section{Preliminaries} \label{background}

\begin{figure*}[t]
\centering
\includegraphics[width=.99\linewidth]{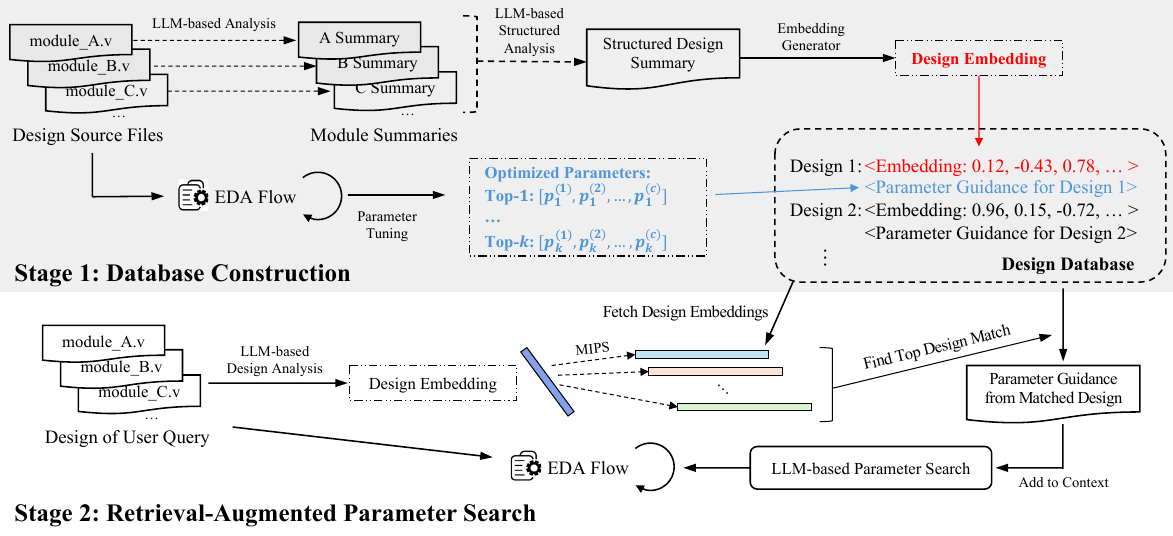}
\caption{Overview of CROP.
In stage 1, CROP constructs a design database. CROP first performs the two-step LLM-based design analysis to produce the design embedding.
CROP selects the top-$k$ parameter sets and their QoR results from pre-existing optimization results to compile the parameter guidance for the circuits in the database.
In stage 2, CROP conducts retrieval-augmented parameter search. First, CROP performs LLM-based analysis on the design of user query and retrieves the parameter guidance of the design from the database with highest similarity based on maximum inner product search (MIPS).
The retrieved parameter guidance is then used to constrain the LLM-based parameter search engine that optimizes the QoR for the input design.
\vspace{-3pt}
}
\label{fig:overview}
\end{figure*}

\subsection{Transformer-based Language Models}

The advent of transformer-based language models has revolutionized the field of natural language processing (NLP) resulting in significant advancements in numerous tasks such as question answering, text generation, and machine translation. Introduced in 2017 by~\cite{vaswani2017attention} the transformer architecture replaced recurrence with multihead self-attention, a mechanism that allowed more efficient training and better performance in sequence modeling tasks. Subsequent developments of larger models such as GPT3~\cite{brown2020gpt3} and LLaMA3~\cite{grattafiori2024llama}, have shown strong zero-shot and few-shot generalization skills when trained on massive corpora. These models are typically pre-trained for next-token generation and later finetuned for downstream tasks such as instruction following or question answering.
Pre-trained transformer-based language models have achieved state-of-the-art accuracy on a wide range of NLP tasks and have shown significant general knowledge capabilities. More recently, new techniques have emerged to improve the reasoning capabilities of LLMs, such as Chain-of-Thought (CoT) prompting and reasoning architectures.
CoT is a prompting technique that encourages step by step reasoning within a traditional LLM and has shown a significant improvement in complex reasoning tasks~\cite{wei2022cot,li2023structuredchainofthoughtpromptingcode}.
Reasoning models are specific architectures designed for reasoning often trained with reinforcement learning and using reasoning tokens to invoke an internal CoT-like process~\cite{openai2024o1}.

\subsection{Retrieval-Augmented Generation}

Retrieval-augmented generation, or RAG, is a popular class of methods to improve the quality of large language model inference by coupling text generation with a context-retrieval mechanism~\cite{lewis2020rag}.
The retrieval mechanism queries external data sources, often databases or collections of documents, to extract relevant information.
This additional information is added to the prompt before text generation is performed, which grounds responses with more up to date and comprehensive information~\cite{petroni2020kilt}. 
A simple retrieval mechanism works by encoding the input query to create an embedding representation which is then used to try to match it to sections of the database that produce similar representations when encoded~\cite{karpukhin2020dense}.
The generator is typically a sequence-to-sequence model (e.g., GPT) which conditions its output on both the input query and the additional retrieved context~\cite{lewis2020rag}.

There are multiple advantages of RAG, including the ability of generative models to produce more factually accurate responses based on retrieved evidence that reduces the susceptibility of models to hallucinations~\cite{lewis2020rag}.
Another major advantage is that the retrieved data allows a model to adapt to domains outside of its training data.
The external database can be updated independently of the model's learned parameters, avoiding the need for expensive retraining or finetuning~\cite{petroni2020kilt}.

\subsection{Parameter Tuning}

Parameter tuning is a critical aspect of optimizing performance in a wide range of computational domains, from machine learning to scientific simulations.
In general, it involves selecting values for a number of configurable parameters that affect the outcome, often trying to optimize for accuracy, execution time, or some other quantifiable metric. 

In the context of EDA, parameter tuning plays a central role in optimizing a design's power usage, area, and performance or timing (PPA).
Modern EDA tools expose a large number of configurable parameters, allowing experienced designers to control aspects of the synthesis, placement, and routing execution.
The parameters include both categorical options, such as Boolean switches, and heuristic approach options, as well as numerical thresholds. In large, modern designs, it has been shown that different sets of parameters frequently lead to drastically different PPA results~\cite{ziegler2016synparam}. 
Traditionally, parameter tuning in EDA tools especially placement is a labor-intensive and time-consuming process relying on the expertise of human designers. Engineers often rely on trial and error approaches in which they iteratively adjust parameters based on the results they produce; this becomes extremely laborious and time-consuming as the EDA tools often take hours or even days to run each iteration~\cite{xie2020fist, agnesina2020vlsi, geng2022ptpt, liang2021flow, xu2024rank}.
Previous work in parameter tuning has shown that there are strong correlations between the most important parameters when optimizing similar designs for specific objectives~\cite{zheng2023boosting, xie2020fist, geng2022ppatuner, zhang2022fastbo}. These works highlight that there are significant benefits to be gained from carrying over information regarding the tuning results between similar designs.

\vspace{6pt}
\section{Problem Formulation} \label{problem_formulation}

We refer to the configurable options in VLSI design flows as parameters.
Each parameter combination is also referred to as a sample or a parameter vector.
A parameter combination $d$ consists of $c$ parameters, and each parameter has $n_i$ options, where $i \in [1, c]$.
Continuous parameters can be discretized into categorical data.
We use $S$ to denote the whole parameter space and $|S| = \prod_{i=1}^{c} n_i$.
The parameter space grows exponentially when $c$ increases.
Due to the large parameter space, limited computational resources, and time constraints, only a small subset $\tilde{S}$ of samples can complete design flows and be evaluated.
The process of selecting samples to form $\tilde{S}$ is referred to as sampling.
The number of trials allowed is denoted as budget $b$, where $|\tilde{S}| \leq b$.

Given an optimization objective, the goal of our parameter tuning framework $F$ is to find the sample with optimal objective value $O$ using no more than $b$ samples.
Assume learning model $f$ is used during exploration, we have:

\begin{equation}
\tilde{S} = F(S, b, f),
\vspace{-10pt}
\end{equation}

\begin{equation}
F^* = \underset{F}{\arg\min}(\min O[\tilde{S}] - \min O[S]).
\end{equation}

To enable efficient parameter space exploration, we define design embeddings $E(D)$ as dense vector representations that capture the semantic and structural characteristics of circuit designs.
These embeddings facilitate similarity-based retrieval from a database of previous designs.
Parameter guidance of a design $D$ is a set of top-$k$ parameter configurations that have shown best performance in terms of the defined optimization objective for this design.

\vspace{6pt}
\section{Methodology of CROP} \label{method}

\subsection{Overview}

CROP is an LLM-powered parameter optimization framework that relies on a circuit knowledge retrieval system.
Figure~\ref{fig:overview} shows the overview of our proposed CROP method, which is composed of two stages.
In stage 1, we construct the database for future knowledge retrieval by the following steps:
1) Process the RTL source files of existing circuits via LLMs to generate module-level analysis summary and then a final summary for the whole circuit;
2) Use an embedding model to generate the embedding of the design based given its final summary as input, and use the design embedding as the indexes in the database.
3) Compile the parameter guidance of the design from pre-existing optimization data.
Specifically, we select the top-$k$ sets of parameter combinations and their QoR values to construct the parameter guidance.
We detail the LLM-based design analysis and embedding generation in subsection~\ref{sec:design-analysis-using-llms}.

In stage 2, CROP is applied to an input design of user query.
The input design is also processed via LLM-based design analysis and design embedding generation.
The generated design embedding is then used to retrieve the parameter guidance of the design with the highest similarity in the database.
The retrieved parameter guidance is then utilized by an LLM-based parameter search engine.
We detail the embedding-based design retrieval method in subsection~\ref{sec:embedding-based-design-retrieval} and the LLM-based parameter search engine in subsection~\ref{sec:rag-enhanced-llm-search}.

\subsection{Design Analysis using LLMs}
\label{sec:design-analysis-using-llms}

The LLM-based design analysis aims to generate a high-level semantic design representation that can be used for a knowledge retrieval system.
First, we prompt an LLM with special instructions to iteratively summarize each module in the RTL code, prompting the model for a structured description of the module; the entries for which we prompt are shown in Table~\ref{tab:module-summary-structure}.
Then, the module-level summaries are concatenated with a new system prompt that queries the model for a structured high-level summary of the circuit's semantic and structural characteristics.
The structure of the high-level summary is shown in Table~\ref{tab:final-summary-structure}.
We prompt the model for a combination of specific traits such as design name and primary inputs/outputs as well as more open-ended features like key design characteristics and potential applications.
This ensures that all of the most important circuit traits are captured as well as using the LLM's prior knowledge to capture less consistent or quantifiable aspects of each design.
This high-level analysis is then passed through an embedding model to retrieve a dense vector representation of the design, which is passed to the RAG engine. 
The iterative summarization of each module is an important first step as it overcomes the scaling issue where the raw RTL code for sufficiently large designs can be tokenized to longer than the LLM's sequence length.
Further discussion of this scaling issue is provided in Section~\ref{discussion}.

\begin{table}[tb]
\centering
\caption{Module-level summary structure in CROP.}
\label{tab:module-summary-structure}
\resizebox{\linewidth}{!}{
\begin{tabular}{p{0.28\linewidth}p{0.68\linewidth}}
\toprule
\textbf{Entry} & \textbf{Description} \\
\midrule
inputs & Hierarchical mapping of input signals for each module, including bit widths and purposes. \\
\midrule
outputs & Hierarchical mapping of output signals for each module, including bit widths and purposes. \\
\midrule
overall functionality & Concise description of what the module does and its primary purpose. \\
\midrule
critical submodules & Analysis of key submodules, their functionality, and internal logic implementation. \\
\midrule
design choices & Architectural decisions made in the implementation, including techniques used. \\
\midrule
timing constraints & Critical paths and timing considerations that affect performance. \\
\bottomrule
\end{tabular}
}
\end{table}

\begin{table}[tb]
\centering
\caption{Final design summary structure in CROP.}
\label{tab:final-summary-structure}
\resizebox{\linewidth}{!}{
\begin{tabular}{p{0.24\linewidth}p{0.72\linewidth}}
\toprule
\textbf{Entry} & \textbf{Description} \\
\midrule
design name & A concise, descriptive name for the overall design based on its functionality. \\
\midrule
design \par functionality & A brief summary of the overall function of the complete circuit. \\
\midrule
primary inputs & Comprehensive list of all main inputs to the design as a whole. \\
\midrule
primary outputs & Comprehensive list of all main outputs from the design as a whole. \\
\midrule
key design \par characteristics & Notable features, architecture choices, and structure in 2-3 sentences. \\
\midrule
key modules and functionalities & Mapping of important modules to their specific functions described concisely. \\
\midrule
module interactions & How the individual modules communicate and interact within the design. \\
\midrule
design optimizations & Notable optimizations in the design that affect performance or efficiency. \\
\midrule
potential \par applications & Possible use cases and applications for this design. \\
\midrule
timing \par considerations & Critical timing paths or constraints that affect overall performance. \\
\bottomrule
\end{tabular}
}
\end{table}

\subsection{Embedding-based Design Retrieval}
\label{sec:embedding-based-design-retrieval}

\begin{figure}[t]
\centering
\includegraphics[width=\linewidth]{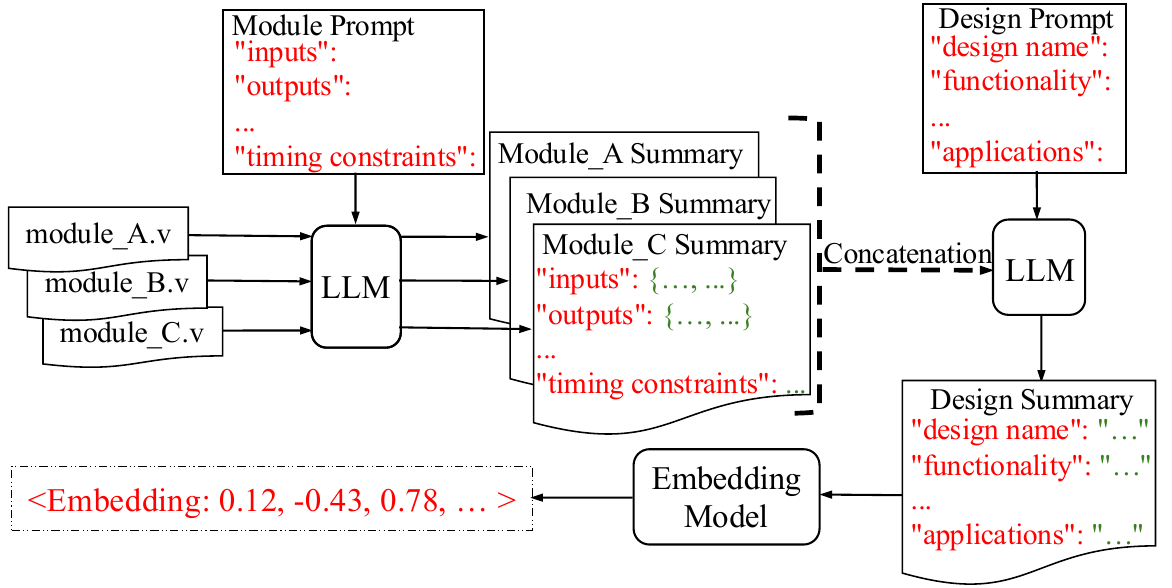}
\caption{Structure for LLM-based design analysis and embedding extraction.}
\label{fig:rtl2spec}
\end{figure}

As shown in Figure~\ref{fig:overview}, CROP constructs a database with embeddings and parameter guidance of pre-existing designs in stage 1.
When a user presents a target design as shown in Stage 2, CROP performs the same LLM-based design analysis process to generate embedding for the target design.
This target design embedding is then compared with each embedding in the database using MIPS, where the design embedding with the highest inner product is chosen.
Since design embeddings represent circuit characteristics in a latent space, when two designs have embeddings with higher inner product values, they likely share more semantic similarities in their functionality and specifications.
This similarity-based retrieval system connects the target design with the most promising parameter guidance, effectively narrowing down the subsequent search process performed by large language models.

\subsection{RAG-enhanced LLM Search}
\label{sec:rag-enhanced-llm-search}

The RAG-enhanced LLM search employs a decoder-only LLM as the parameter search engine.
During each iteration, CROP constructs 
a prompt using general information, the parameter names and valid options, data retrieved through the RAG module, and the historical results of previous iterations.
This prompt is provided to the model through an API which returns a response that is parsed for the suggested parameter sample.
This sample is then validated and evaluated using RTL Architect, and the result is included in the prompts of future iterations. 
The general information consists of background and instructions.
The background explains that the model is impersonating an experienced EDA engineer and tuning EDA flow parameters for QoR objectives.
The instructions outline the desired response format, as well as informing the model to be aware of both exploration and exploitation.
We use a CoT model, as they have been shown to excel at tasks that require advanced scientific reasoning and domain-specific knowledge \cite{cui2025curie, openai2024o1}. 
Additionally, following \cite{li2023structuredchainofthoughtpromptingcode}, we construct our prompts with clear instructions which makes the CoT model adhere to a consistent output format.
This process allows an LLM to act as a fully autonomous design engineer leveraging prior knowledge while thoroughly exploring the search space, freeing human engineers from this laborious task. 

\vspace{6pt}
\section{Experiment Results} \label{experiments}

\subsection{Experiment Setup}

We evaluate CROP using a VLSI design flow based on Synopsys RTL Architect.
We optimize 31 parameters related to placement optimization.
Examples of these parameters are shown in Table~\ref{tab:parameters}.
Each parameter corresponds to a configurable option of the RTL Architect flow.
The parameters are selected to target coarse placement in the VLSI flow and are categorized into four groups for clarity: density, congestion, timing, and power.
The parameters we use were selected based on industry best practices and have shown substantial performance improvements.
Each parameter has a list of valid values for optimization algorithms to select, either as a categorical parameter or a continuous one which has been discretized into a set number of options.
This gives us a fully discrete search space with more than $2.8\times10^{15}$ possibilities. The size of the search space combined with the runtime of commercial EDA tools makes exhaustive search impossible.
We employ the RTLRewriter benchmark~\cite{yao2024rtlrewriter} and two proprietary industrial processor designs to evaluate our framework.
The RTLRewriter benchmark consists of a diverse collection of open-sourced RTL designs spanning various complexity levels and application domains.
They provide a diverse range of designs on which we can evaluate the efficacy of our RAG engine.
Additionally, we use two proprietary processor designs, which we refer to as \textit{Prior Core} and \textit{Target Core} to test the efficacy of CROP's ability to transfer knowledge to new designs.
Both cores consist of 111K cells and share similar architectural characteristics, but are implemented in different technology nodes, specifically, a 7 nm industrial tech node for \textit{Prior Core} and 5 nm for \textit{Target Core}.

We implemented three state-of-the-art baseline approaches for comparison on the same task: random search, Optuna, and Bayesian optimization (BO).
As a naive baseline, random search selects each parameter value independently and without regard to previous outcomes.
In each iteration, a new configuration is generated by randomly selecting an index for each parameter from its list of options using Python's built-in random module, resulting in a uniformly random distribution.
Each trial is sampled i.i.d. from the search space, so this method does not utilize any historical information from previous evaluations.
As another baseline, we used the Optuna hyperparameter optimization framework~\cite{akiba2019optuna} to perform a guided search over the discrete search space.
In our implementation,
each of the 31 parameters is treated as a categorical variable and the optimizer makes a selection for each using \texttt{trial.suggest\_categorical} which samples from the predefined set of values.
We used Optuna's default sampling strategy, tree-structured parzen estimator, to control the search and manage the exploration-exploitation trade-off.
The final baseline uses a Gaussian process-based optimizer implemented via the BayesianOptimization Python package~\cite{garrido2020dealing}.
This process models the optimization problem as a black box and fits a Gaussian process to the known points.
Then it uses the Expected Improvement exploration strategy to select the next point of exploration~\cite{nogueira2014bo}.
We initialize this optimizer with ten points randomly sampled from the search space before commencing the Gaussian process, this ensures sufficient initial exploration.
Throughout all experiments with CROP, we use an instance of GPT-4o for the design analysis, all-MiniLM-L6-v2 for text embedding of the design summary, and OpenAI o1 for the LLM-based search.


\begin{table}[t]
\centering
\caption{Flow parameter examples for RTL Architect.}
\label{tab:parameters}
\resizebox{\linewidth}{!}{
\begin{tabular}{cll}
\toprule
\textbf{Category} & \textbf{Parameter} & \textbf{Value Range} \\
\midrule
\multirow{6}{*}{Density}
 & density\_control\_version & \{1, 2, 3\} \\
 & enable\_pin\_density\_aware & \{true, false\} \\
 & enable\_ultra\_effort & \{true, false\} \\
 & max\_density & (0.0 - 1.0): 5 values \\
 & max\_density\_adjustment & (0.0 - 1.0): 5 values \\
 & max\_pins\_density & (1.0 - 1000000.0): 7 values \\
\midrule
\multirow{6}{*}{Congestion}
 & congestion\_level & (0.0 - 5.0): 10 values \\
 & congestion\_mode & \{1, 2\} \\
 & congestion\_style & \{1, 2\} \\
 & congestion\_version & \{1, 2\} \\
 & enable\_blockage\_aware & \{true, false\} \\
 & enable\_congestion\_aware & \{true, false\} \\
\midrule
\multirow{11}{*}{Timing}
 & buffer\_aware\_mode & \{0, 1, 2, 3, 4\} \\
 & clock\_weight & \{0, 1, 2, 3\} \\
 & delay\_weight & (0.0 - 1.0): 5 values \\
 & enable\_buffer\_aware & \{true, false\} \\
 & enable\_clock\_aware & \{true, false\} \\
 & enable\_clock\_latency\_aware & \{true, false\} \\
 & enable\_clockgate\_aware & \{true, false\} \\
 & enable\_clockgate\_identification & \{true, false\} \\
 & enable\_clockgate\_latency\_aware & \{true, false\} \\
 & enable\_path\_aware & \{true, false\} \\
 & enable\_registers\_aware & \{true, false\} \\
 & max\_delay\_weight & (1.0 - $10^{20}$): 11 values \\
 \midrule
 \multirow{3}{*}{Power}
 & power\_mode & \{0, 1, 2, 3, 4\} \\
 & power\_weight & (0.0 - 1.0): 5 values \\
 & power\_weight\_stdev &  (0.0 - 100.0): 10 values \\
\bottomrule
\end{tabular}
}
\vspace{-10pt}
\end{table}

\subsection{Circuit Retrieval Quality}

The efficacy of a RAG framework for VLSI circuit optimization is inherently contingent upon the precision and relevance of the retrieval mechanism.
The following experiments present an evaluation of our retrieval component, which constitutes the foundational element of the proposed framework.
We quantitatively assess the retrieval quality through its retrieval precision, which measures the semantic alignment between query circuits and retrieved exemplars.

\begin{figure}[b]
\centering
\includegraphics[width=.95\linewidth]{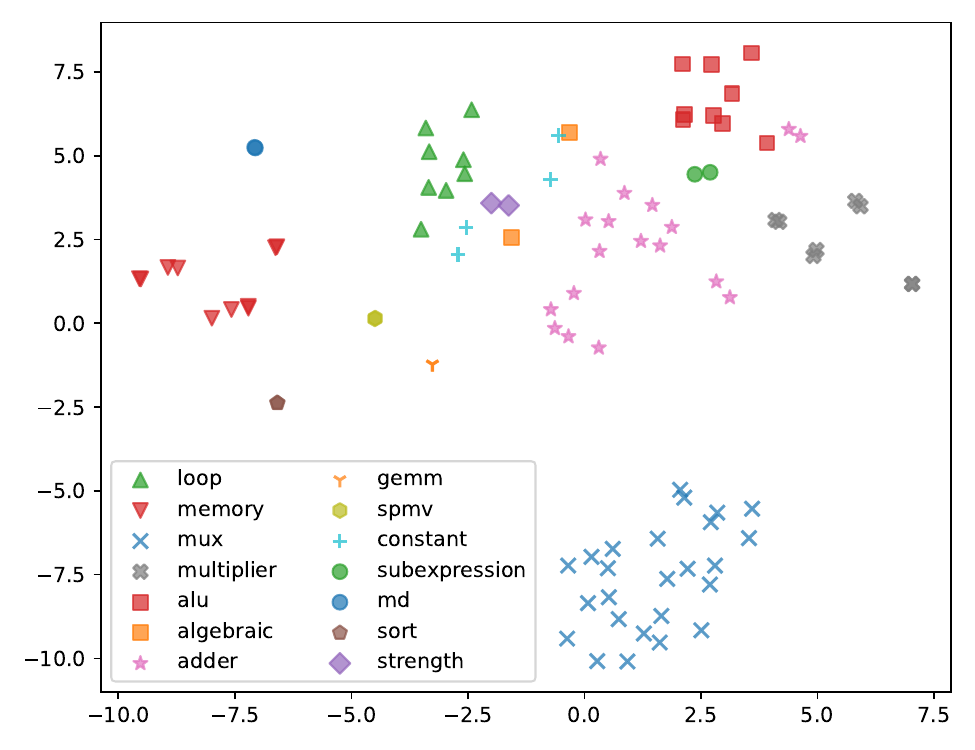}
\caption{A t-SNE-based visualization of design description embeddings colored by each design's category, designs from the RTLRewriter benchmark.
}
\label{fig:circuit-embeddings}
\end{figure}

Figure~\ref{fig:circuit-embeddings} shows a t-SNE visualization of design description embeddings colored by each design's category, where individual points represent RTL module embedding vectors displayed with different colors and markers to indicate distinct design categories.
The embeddings, generated from module source code and projected into 2D space using t-SNE, reveal clustering patterns that highlight both semantic similarities between designs within the same category and interesting cross-category relationships. For example, the alu, adder, multiplier, and subexpression circuits are all closely clustered, highlighting their shared relation to mathematical calculation (the subexpression circuits both implement a bitwise multiplier). Table~\ref{tab:retrieval_precision} shows the precision scores for CROP's circuit retrieval mechanism on different families of designs. These results show an average precision score of 0.906 at $k=1$ and 0.819 at $k=3$, once again highlighting the ability of the design retrieval engine to successfully match designs with those that are semantically or functionally similar.

\begin{table}[tb]
\centering
\caption{Retrieval performance evaluation of circuit embeddings.}
\label{tab:retrieval_precision}
\begin{subtable}{\linewidth}
\centering
\caption{Average Precision and Recall at different $k$ values}
\begin{tabular}{lcccccc}
\toprule
\textbf{Metric @k} & \textbf{@1} & \textbf{@3} & \textbf{@5} & \textbf{@10} & \textbf{@20} \\
\midrule
Precision & 0.906 & 0.819 & 0.767 & 0.651 & 0.506 \\
Recall & 0.193 & 0.331 & 0.441 & 0.636 & 0.852 \\
\bottomrule
\end{tabular}
\end{subtable}

\vspace{0.3cm}

\begin{subtable}{\linewidth}
\centering
\caption{Precision@1 by Circuit Category}
\begin{tabular}{lclc}
\toprule
\textbf{Category} & \textbf{Precision@1} & \textbf{Category} & \textbf{Precision@1} \\
\midrule
mux & 1.000 & md & 1.000 \\
memory & 1.000 & spmv & 1.000 \\
multiplier & 1.000 & alu & 0.889 \\
subexpression & 1.000 & adder & 0.889 \\
gemm & 1.000 & loop & 0.875 \\
strength & 1.000 & constant & 0.250 \\
sort & 1.000 & algebraic & 0.000 \\
\bottomrule
\end{tabular}
\end{subtable}

\vspace{0.3cm}

\begin{subtable}{\linewidth}
\centering
\caption{Dataset Statistics: 96 embeddings across 14 categories}
\label{tab:dataset_stats}
\begin{tabular}{lclc}
\toprule
\textbf{Category} & \textbf{Count} & \textbf{Category} & \textbf{Count} \\
\midrule
mux & 25 & memory & 10 \\
adder & 18 & multiplier & 8 \\
alu & 9 & loop & 8 \\
constant & 4 & algebraic & 2 \\
gemm & 2 & md & 2 \\
sort & 2 & spmv & 2 \\
strength & 2 & subexpression & 2 \\
\bottomrule
\end{tabular}
\end{subtable}
\end{table}

\subsection{Parameter Tuning Performance}

\begin{table}[tb]
\centering
\caption{Comparison of \textit{Target Core}'s total power consumption (mW) using different parameter optimization algorithms.}
\begin{tabular}{lcccc}
\toprule
\textbf{Method} & \textbf{5 Iterations} & \textbf{10} & \textbf{50} & \textbf{75} \\
\midrule
\multicolumn{1}{l|}{\textbf{Optuna}} & 134.63 & 134.63 & 133.45 & 133.24 \\
\multicolumn{1}{l|}{\textbf{BO}} & 137.72 & 134.51 & 133.68 & 133.17 \\
\multicolumn{1}{l|}{\textbf{Random Search}} & 135.52 & 135.11 & 133.44 & 133.44 \\
\multicolumn{1}{l|}{\textbf{CROP}} & \textbf{127.08} & \textbf{127.08} & \textbf{121.01} & \textbf{119.98} \\
\bottomrule
\end{tabular}
\label{tab:coreb_results}
\end{table}
In these experiments we report power as the primary metric being optimized however, at the end of this section we present an analysis of the tradeoff between different PPA metrics.
Figure~\ref{fig:opt-result_B} compares the performance of CROP against random search, BO and Optuna on \textit{Target Core}.
The results shown in Table~\ref{tab:coreb_results} demonstrate that CROP consistently outperforms all baseline methods across various search budget sizes.
Specifically, CROP achieves a clear advantage over the baselines reaching 127.08 mW after only 5 iterations, and converges to a minimal power of 119.98 mW throughout the experiment.
This accelerated convergence behavior suggests that the context enrichment from retrieving similar circuits (i.e., \textit{Prior Core}) provides valuable guidance to effectively constrain the parameter search process.
On the other hand, the lowest total power achieved by all other baselines after 75 iterations is 133.17 mW, which means that CROP achieved a \textbf{9.9\%} reduction of the total power consumption compared to all baseline approaches.
This substantial improvement highlights the efficacy of knowledge transfer between similar designs implemented in different technology nodes, as well as the LLMs ability to guide efficient search.
Notably, while traditional optimization methods continue to make incremental improvements throughout their search trajectories, they fail to identify the high-quality parameter regions that CROP discovers almost immediately through its retrieval-augmented approach, demonstrating the limitations of design-agnostic search strategies for modern VLSI designs.

\begin{figure}[b]
\centering
\includegraphics[width=\linewidth]{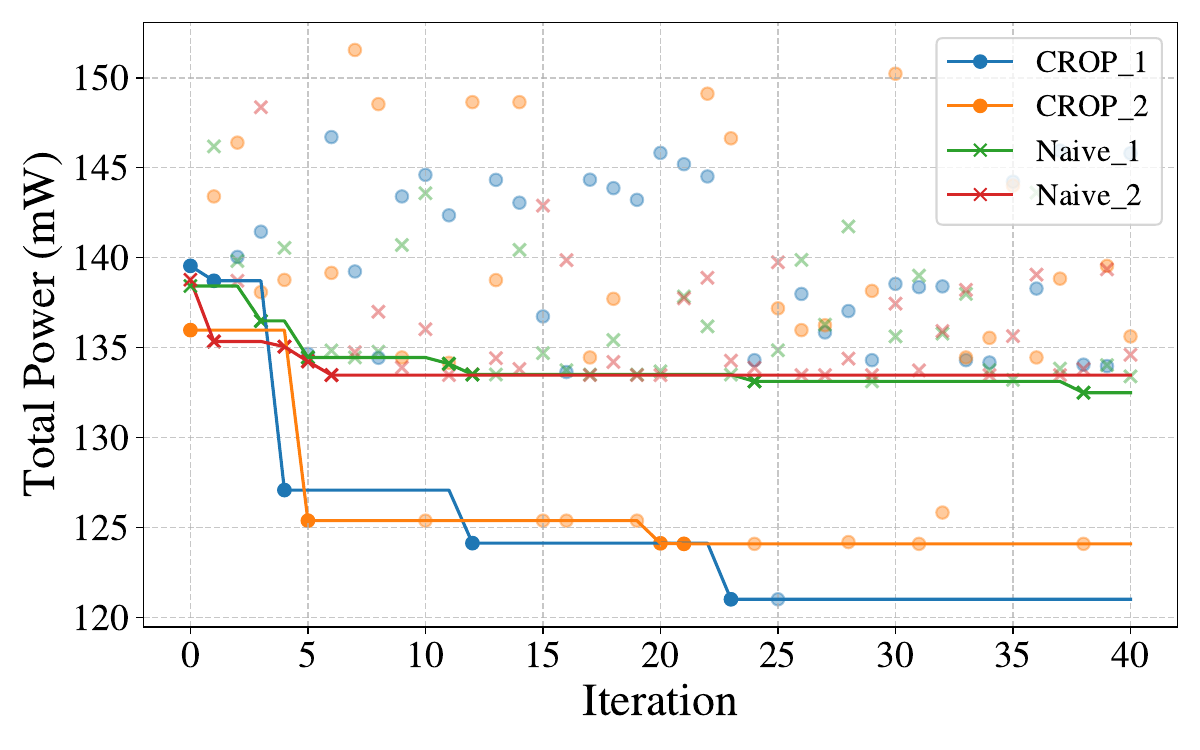}
\caption{Comparison of CROP and naive LLM search (without RAG-based parameter guidance retrieval) across two independent runs each. 
}
\label{fig:rag_vs_naive}
\end{figure}

\subsection{RAG Benefit Analysis}
\begin{figure}[tb]
\centering
\includegraphics[width=\linewidth]{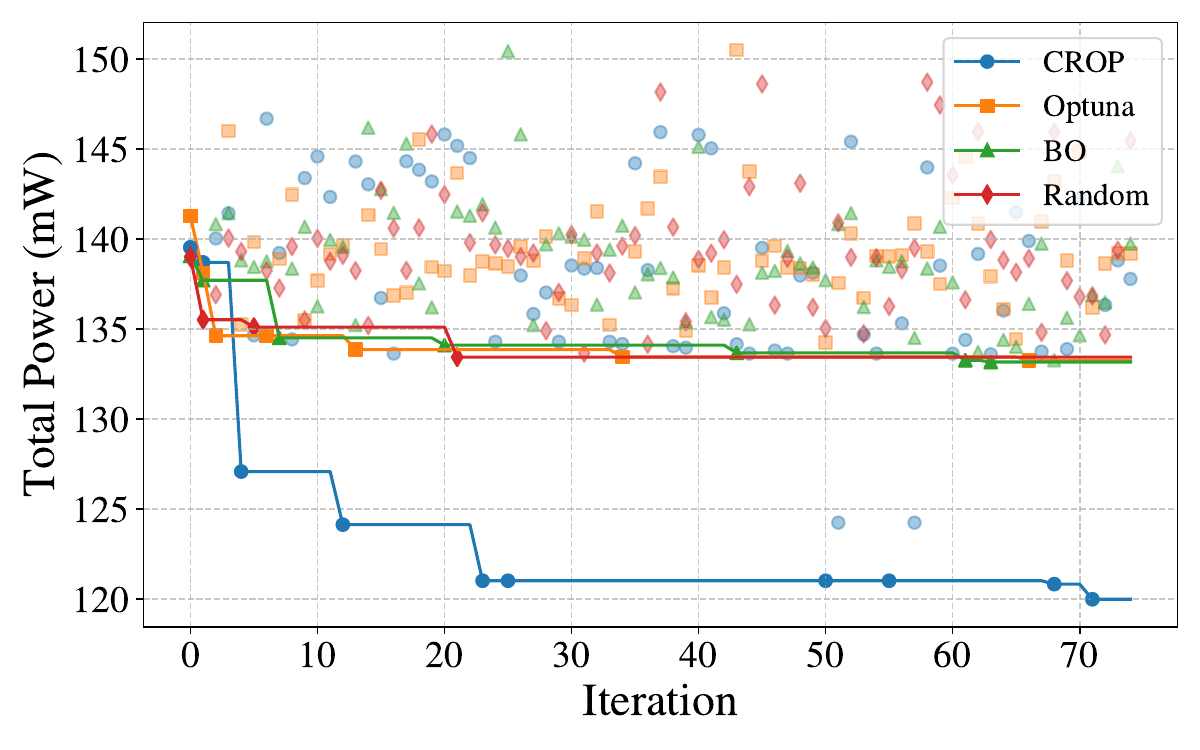}
\caption{Comparing the total power usage found by CROP and the baseline methods on \textit{Target Core}.
\vspace{-10pt}
}
\label{fig:opt-result_B}
\end{figure}

In order to evaluate the efficacy of the RAG component within CROP we compare CROP to a naive LLM-based search using the same model prompted without the information from the design retrieval component.
We ran multiple iterations of both searches, which are shown in Figure~\ref{fig:rag_vs_naive}.
Including RAG information in the search prompting resulted in a \textbf{6.59\%} and \textbf{8.66\%} reduction in power usage after iterations 5 and 40, respectively, compared to the best values from naive LLM search traces.
The performance of the naive LLM is comparable to those of the baseline methods shown in Figure~\ref{fig:opt-result_B} showing that LLM guided search on its own is a valid method for parameter tuning.
This relies on the LLM's ability to intuitively understand the impact of parameters, tracking how they effect the result over time using only the knowledge encoded in the LLM's training. 
However, the RAG-aware LLM significantly outperforms any other method. 
In CROP, RAG enables faster convergence to optimal values, achieving superior results with fewer iterations across all trials.
These results highlight both the efficacy of the RAG component in CROP's search and the ability of CROP to save significant amounts of time and computational resources.

\subsection{Parameter Importance Analysis}

\begin{figure}[b]
\centering
\includegraphics[width=\linewidth]{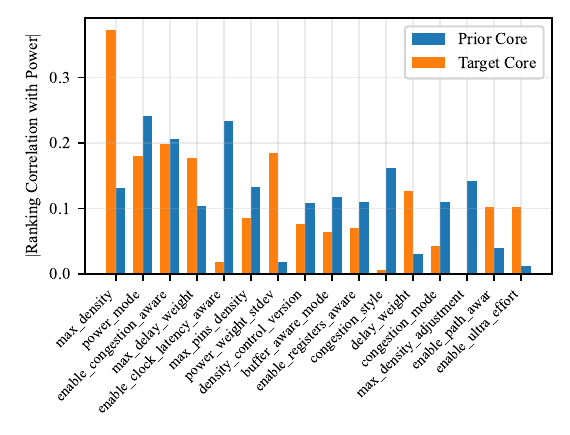}
\caption{
The absolute Spearman's rank correlation between parameters and power consumption across target and prior processor cores.
}
\label{fig:parameter-effects}
\end{figure}

Spearman's rank correlation coefficient~\cite{spearman1961proof} measures the strength and direction of monotonic relationships between two variables based on the ranked values rather than raw values.
The coefficient $\rho$ is calculated as $\rho=1-\frac{6\sum d_i^2}{n(n^2)-1}$, where $d_i$ is the difference between the ranks of corresponding values and $n$ is the number of value pairs.
This metric quantifies parameter importance by assessing the monotonic association between parameter values and performance metrics without assuming linearity.
Figure~\ref{fig:parameter-effects} compares the absolute Spearman's rank correlation coefficients between the top-impact parameters ($\left|\rho\right| > 0.10$) and power consumption across the two processor cores.
We use absolute values to focus on the magnitude of each parameter's influence on power (i.e., its importance in terms of power optimization).
Based on the absolute Spearman's ranking correlation data, we observe that high-importance parameters tend to be influential across both cores, suggesting underlying commonalities in design sensitivities.
Although, some core-specific variations do exist such as \texttt{power\_weight\_stdev} (0.185, 0.018) which significantly impacts \textit{Target Core} but not \textit{Prior Core}, while \texttt{enable\_clock\_latency\_aware} (0.018, 0.234) and \texttt{congestion\_style} (0.006, 0.161) show the opposite pattern.  
The most influential parameters demonstrate strong importance across both cores, including \texttt{max\_density} (0.372, 0.131), \texttt{power\_mode} (0.180, 0.242), and \texttt{enable\_congestion\_aware} (0.199, 0.206).
These are highly important parameters regardless of the current technology node. This supports the idea that optimized parameter sets from previous designs can be used to quickly understand the most important directions to explore within the search space accelerating convergence.

\subsection{Runtime Analysis}

\begin{table}[tb]
\centering
\caption{Runtime breakdown by framework component for optimizers on \textit{Target Core}.}
\label{tab:runtime_analysis}
\begin{subtable}{\linewidth}
\centering
\caption{CROP runtime analysis.}
\begin{tabular}{lccc}
\toprule
\textbf{Component} & \textbf{1 Iteration} & \textbf{10 Iterations} & \textbf{50 Iterations} \\
\midrule
RAG Overhead        & 7.18\%  & 0.77\%  & 0.15\% \\
Parameter Selection & 0.99\%  & 1.05\%  & 1.06\% \\
EDA Flow            & 91.83\% & 98.18\% & 98.79\% \\
\bottomrule
\end{tabular}
\end{subtable}

\vspace{1em}

\begin{subtable}{\linewidth}
\centering
\caption{BO runtime analysis.}
\begin{tabular}{lccc}
\toprule
\textbf{Component} & \textbf{1 Iteration} & \textbf{10 Iterations} & \textbf{50 Iterations} \\
\midrule
Parameter Selection & 0.14\%  & 0.14\%  & 0.14\% \\
EDA Flow            & 99.86\% & 99.86\% & 99.86\% \\
\bottomrule
\end{tabular}
\end{subtable}

\vspace{1em}

\begin{subtable}{\linewidth}
\centering
\caption{Optuna runtime analysis.}
\begin{tabular}{lccc}
\toprule
\textbf{Component} & \textbf{1 Iteration} & \textbf{10 Iterations} & \textbf{50 Iterations} \\
\midrule
Parameter Selection & 0.15\%  & 0.15\%  & 0.15\% \\
EDA Flow            & 99.85\% & 99.85\% & 99.85\% \\
\bottomrule
\end{tabular}
\end{subtable}
\end{table}

As the overall goal of this work is to reduce the time parameter tuning takes, it is important to evaluate the runtime of our approach.
Table~\ref{tab:runtime_analysis} shows the runtime breakdown for CROP as well as BO and Optuna.
The runtime analysis of each of these methods reveals that running the EDA flow dominates the total runtime of the experiments.
The runtime analysis reveals that running the EDA flow dominates the total runtime across all methods, leading to relatively marginal differences between approaches.

However, several distinctions emerge in the initialization and iteration phases.
CROP incurs additional initialization overhead from the RTL to embedding extraction and similarity-base retrieval mechanism, while the baselines have negligible initialization costs.
Nevertheless, CROP's initialization overhead is amortized across subsequent iterations, reducing its relative impact as more iterations are run.
The methods also differ in how they spend time during each iteration.
CROP's parameter selection stage is mostly spent waiting on the inference of the reasoning model, whereas Optuna and BO's are dominated by logging and file cleanup, with the actual next parameter calculation requiring a small fraction of that time.

While automated approaches will always outperform human experts on a per-iteration basis, CROP's key advantage lies in solution quality rather than speed per iteration.
CROP's ability to find notably better solutions within significantly fewer iterations allows for the potential of early stopping leading to a massive save in overall runtime compared to BO and Optuna.

\subsection{PPA-tradeoff analysis}

\begin{figure}[tb]

\begin{subfigure}{\linewidth}
\includegraphics[width=\linewidth]{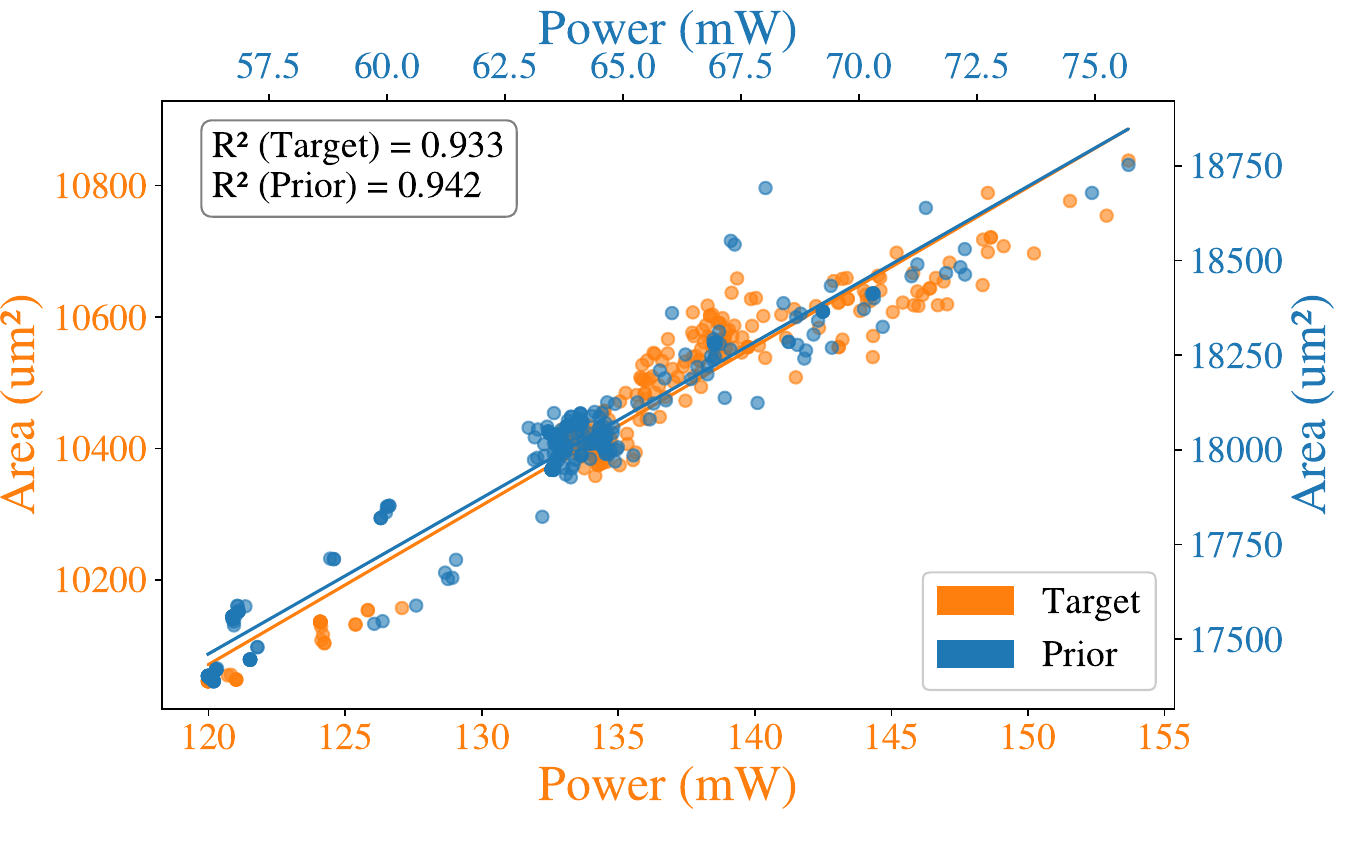}
\caption{
Correlation analysis between power and area.
\vspace{6pt}
}
\label{fig:power_vs_area}
\end{subfigure}

\begin{subfigure}{\linewidth}  
\includegraphics[width=\linewidth]{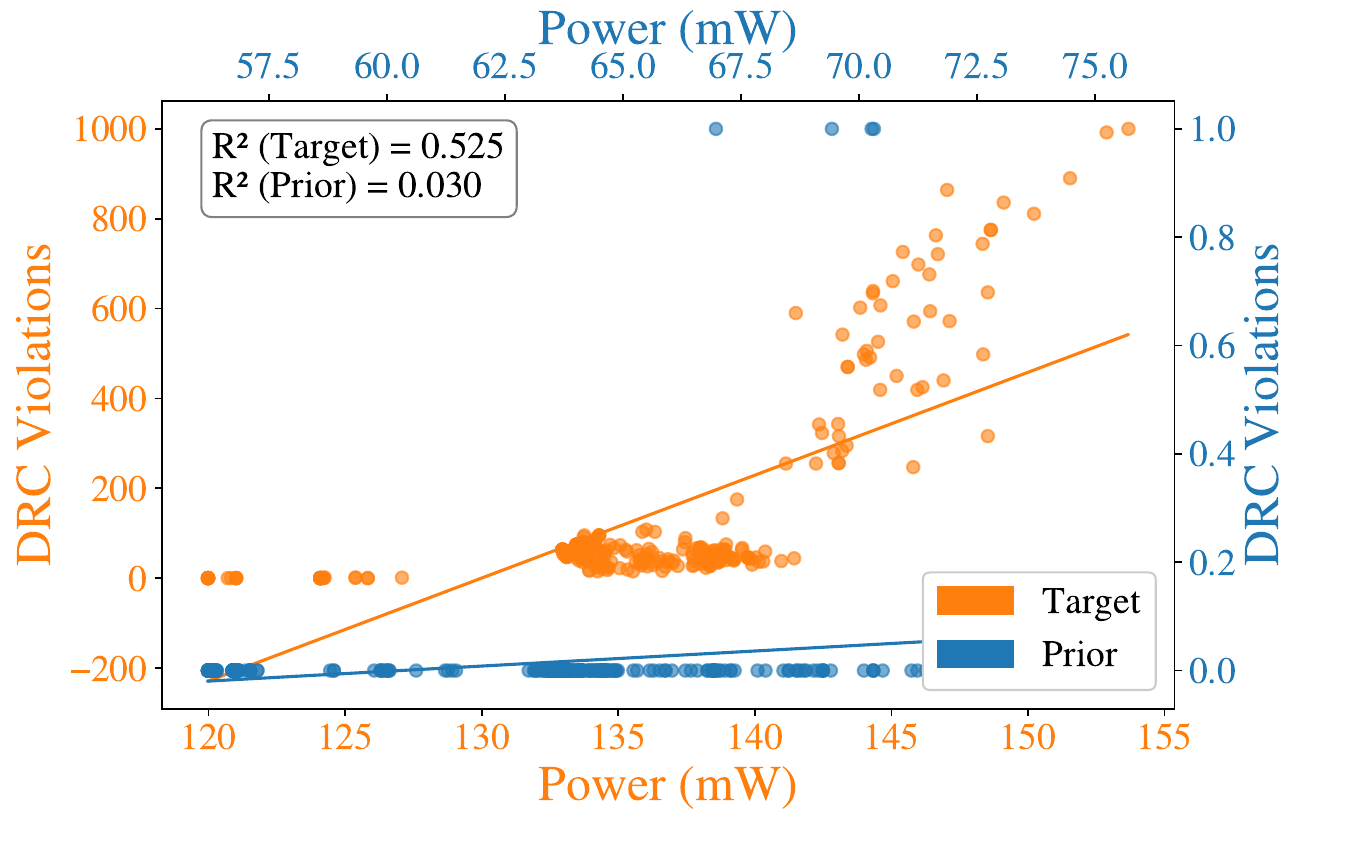}
\caption{
Correlation analysis between power and number of DRC violations.
\vspace{6pt}
}
\label{fig:power_vs_drc}
\end{subfigure}

\begin{subfigure}{\linewidth}
\includegraphics[width=\linewidth]{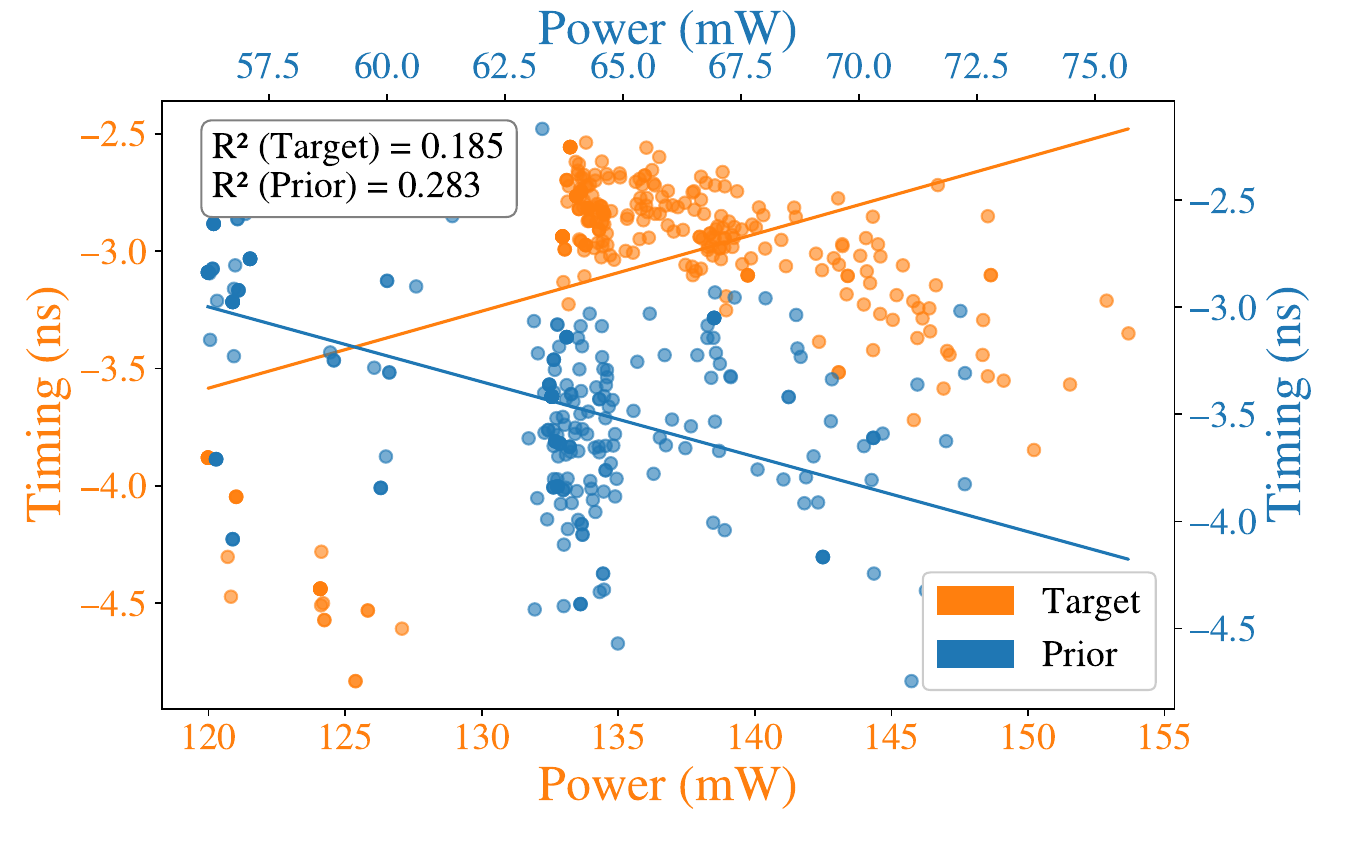}
\caption{
Correlation analysis between power and timing in terms of negative slack.
\vspace{6pt}
}
\label{fig:power_vs_timing}
\end{subfigure}

\caption{
Correlation analysis between power and other QoR values for both \textit{Prior Core} and \textit{Target Core}. 
}
\end{figure}

The primary focus of the experiments presented early in this paper is power optimization.
VLSI design is usually considered a multi-objective optimization problem where the design's power, area, timing, and number of design rule violations are all important.
We thoroughly analyzed the correlation between these different values using the output of RTL Architect for both \textit{Prior Core} and \textit{Target Core} and concluded that power was an appropriate priority.
Figure~\ref{fig:power_vs_area} shows that the designs' power and area results have a very strong positive correlation, meaning that the results with low power usage also have a small area. 
Figure~\ref{fig:power_vs_drc} shows a positive correlation between the number of DRC violations and power values, but more importantly, all the lowest power values found have zero design rule violations. Overall \textit{Prior Core} has far fewer DRC violations as a result of the fact that the newer technology nodes have more design rules to adhere to.
When analyzing timing (in terms of total negative slack), we found a generally weak correlation between timing and power shown in Figure~\ref{fig:power_vs_timing}. However, it should be noted that the LLM guided search on \textit{Target Core} sacrificed timing to optimize power and area, this is shown by the orange dots in the bottom left corner. The lowest power point found by the search had a total of 3.88 ns of negative slack, while the most timing-aware points had only 2.54 ns of negative slack.

\vspace{6pt}
\section{Discussion} \label{discussion}

\paragraph{\textbf{Addressing LLM scalability issue}}
The scalability of LLMs when processing RTL code is a fundamental challenge that our method addresses through an iterative summarization approach.
The context window size of LLMs is crucial to their ability to craft coherent, accurate, and relevant responses for code analysis~\cite{li2023loogle,xu2022systematic,liu2024lost}.
Industrial RTL codebases frequently exceed tens of thousands of lines, creating significant challenges for direct processing in LLMs due to the quadratic computational complexity of transformer attention mechanisms.

Our iterative summarization approach in Figure~\ref{fig:rtl2spec} overcomes this limitation by first processing individual modules separately and then synthesizing these summaries into a coherent design representation.
This two-stage process effectively compresses the semantic content of large RTL codebases while preserving their functional characteristics.
By generating embeddings from these structured summaries rather than raw RTL files, we maintain the essential design characteristics while dramatically reducing the token count required for processing.
This approach also standardizes the representation format, making similarity comparisons more reliable across designs of varying complexity and size.

This method proves particularly valuable for industrial-scale designs where direct embedding of the complete RTL code could easily exceed even the expanded context windows of modern LLMs (e.g., 128K tokens for GPT-4o in our experimental setting).
Our experiments confirm that these derived embeddings successfully capture semantic similarities between functionally related circuits, as evidenced by the clustering patterns in Figure~\ref{fig:circuit-embeddings}, while avoiding the prohibitive computational costs of processing entire RTL repositories directly.

\paragraph{\textbf{Future directions}}
The experimental evaluations of CROP on industrial designs show that it is a promising direction for improving automated flow tuning.
We identify the following areas for future improvement:
1) The inherent database reliance of CROP presents both strengths and challenges;
while it demonstrates the strength of achieving better QoR results in significantly fewer iterations as shown in our evaluations, the challenge emerges as the database expands and may require a more sophisticated method for design retrieval that maintains accuracy and efficiency.
2) This retrieval-driven approach is complemented by CROP's tool-agnostic nature, allowing seamless integration with existing EDA toolchains and compatibility with approaches that use jump-start or early-stopping methods.
Methods such as REMOTune~\cite{zheng2023boosting} and FlowTuner~\cite{liang2021flow} that reduce EDA tool runtime can be directly incorporated, potentially further improving CROP's overall efficiency as demonstrated in our runtime analysis.
3) Significant performance gains can be achieved through parallelization, as CROP can be combined with search parallelization methods such as separate trust regions.
This represents a highly beneficial improvement direction since database results can reduce initial iterations required to establish trust regions, which can then be searched in parallel using multiple LLM-guided search instances. 
4) CROP's model-agnostic design enables improvements through enhanced LLMs,
whether by adopting newer and more capable models,
implementing fine-tuned domain-specific models,
or applying training-free decoding techniques that reduce hallucination~\cite{zhang2025sledselflogitsevolution, obrien2023contrastivedecodingimprovesreasoning}.

\vspace{6pt}
\section{Conclusion} \label{conclusion}

In this paper, we propose CROP, a novel framework for EDA tool parameter tuning based on the ability of an LLM to take the place of an experienced design engineer using prior knowledge extracted from a database of existing designs. We introduce a novel method for finding semantically similar designs using a dense vector representation of a design as well as a guided LLM based search, which is able to autonomously transfer prior knowledge from previous experiments. Our experimental results on industrial designs highlight that CROP is able to find better results in fewer iterations than existing approaches. Finally, future work is available combining CROP with existing methods to improve the runtime of EDA tools through approaches such as early stopping or parallelization.

\vspace{6pt}
\section*{Acknowledgment}

This work is in part supported by NSF 2106828 and Synopsys gift.

\bibliographystyle{IEEEtran}
\bibliography{references}

\begin{thebibliography}{10}
\providecommand{\url}[1]{#1}
\csname url@samestyle\endcsname
\providecommand{\newblock}{\relax}
\providecommand{\bibinfo}[2]{#2}
\providecommand{\BIBentrySTDinterwordspacing}{\spaceskip=0pt\relax}
\providecommand{\BIBentryALTinterwordstretchfactor}{4}
\providecommand{\BIBentryALTinterwordspacing}{\spaceskip=\fontdimen2\font plus
\BIBentryALTinterwordstretchfactor\fontdimen3\font minus \fontdimen4\font\relax}
\providecommand{\BIBforeignlanguage}[2]{{%
\expandafter\ifx\csname l@#1\endcsname\relax
\typeout{** WARNING: IEEEtran.bst: No hyphenation pattern has been}%
\typeout{** loaded for the language `#1'. Using the pattern for}%
\typeout{** the default language instead.}%
\else
\language=\csname l@#1\endcsname
\fi
#2}}
\providecommand{\BIBdecl}{\relax}
\BIBdecl

\bibitem{geng2022ptpt}
H.~Geng, T.~Chen, Y.~Ma, B.~Zhu, and B.~Yu, ``Ptpt: Physical design tool parameter tuning via multi-objective bayesian optimization,'' \emph{IEEE transactions on computer-aided design of integrated circuits and systems}, vol.~42, no.~1, pp. 178--189, 2022.

\bibitem{xie2020fist}
Z.~Xie, G.-Q. Fang, Y.-H. Huang, H.~Ren, Y.~Zhang, B.~Khailany, S.-Y. Fang, J.~Hu, Y.~Chen, and E.~C. Barboza, ``Fist: A feature-importance sampling and tree-based method for automatic design flow parameter tuning,'' in \emph{2020 25th Asia and South Pacific Design Automation Conference (ASP-DAC)}.\hskip 1em plus 0.5em minus 0.4em\relax IEEE, 2020, pp. 19--25.

\bibitem{geng2022ppatuner}
\BIBentryALTinterwordspacing
H.~Geng, Q.~Xu, T.-Y. Ho, and B.~Yu, ``Ppatuner: pareto-driven tool parameter auto-tuning in physical design via gaussian process transfer learning,'' in \emph{Proceedings of the 59th ACM/IEEE Design Automation Conference}, ser. DAC '22.\hskip 1em plus 0.5em minus 0.4em\relax New York, NY, USA: Association for Computing Machinery, 2022, p. 1237–1242. [Online]. Available: \url{https://doi.org/10.1145/3489517.3530602}
\BIBentrySTDinterwordspacing

\bibitem{zheng2023boosting}
S.~Zheng, H.~Geng, C.~Bai, B.~Yu, and M.~D. Wong, ``Boosting vlsi design flow parameter tuning with random embedding and multi-objective trust-region bayesian optimization,'' \emph{ACM Transactions on Design Automation of Electronic Systems}, vol.~28, no.~5, pp. 1--23, 2023.

\bibitem{ziegler2016synparam}
M.~M. Ziegler, H.-Y. Liu, G.~Gristede, B.~Owens, R.~Nigaglioni, and L.~P. Carloni, ``A synthesis-parameter tuning system for autonomous design-space exploration,'' in \emph{2016 Design, Automation \& Test in Europe Conference \& Exhibition (DATE)}, 2016, pp. 1148--1151.

\bibitem{ma2019cad}
Y.~Ma, Z.~Yu, and B.~Yu, ``Cad tool design space exploration via bayesian optimization,'' in \emph{2019 ACM/IEEE 1st Workshop on Machine Learning for CAD (MLCAD)}.\hskip 1em plus 0.5em minus 0.4em\relax IEEE, 2019, pp. 1--6.

\bibitem{akiba2019optuna}
T.~Akiba, S.~Sano, T.~Yanase, T.~Ohta, and M.~Koyama, ``Optuna: A next-generation hyperparameter optimization framework,'' in \emph{Proceedings of the 25th ACM SIGKDD international conference on knowledge discovery \& data mining}, 2019, pp. 2623--2631.

\bibitem{kwon2019learning}
J.~Kwon, M.~M. Ziegler, and L.~P. Carloni, ``A learning-based recommender system for autotuning design flows of industrial high-performance processors,'' in \emph{Proceedings of the 56th Annual Design Automation Conference 2019}, 2019, pp. 1--6.

\bibitem{agnesina2020vlsi}
A.~Agnesina, K.~Chang, and S.~K. Lim, ``Vlsi placement parameter optimization using deep reinforcement learning,'' in \emph{Proceedings of the 39th international conference on computer-aided design}, 2020, pp. 1--9.

\bibitem{liang2021flow}
R.~Liang, J.~Jung, H.~Xiang, L.~Reddy, A.~Lvov, J.~Hu, and G.-J. Nam, ``Flowtuner: A multi-stage eda flow tuner exploiting parameter knowledge transfer,'' in \emph{2021 IEEE/ACM International Conference On Computer Aided Design (ICCAD)}, 2021, pp. 1--9.

\bibitem{hsiao2024fasttuner}
H.-H. Hsiao, Y.-C. Lu, P.~Vanna-Iampikul, and S.~K. Lim, ``Fasttuner: Transferable physical design parameter optimization using fast reinforcement learning,'' in \emph{Proceedings of the 2024 International Symposium on Physical Design}, 2024, pp. 93--101.

\bibitem{xu2024rank}
P.~Xu, S.~Zheng, Y.~Ye, C.~Bai, S.~Xu, H.~Geng, T.-Y. Ho, and B.~Yu, ``Ranktuner: When design tool parameter tuning meets preference bayesian optimization,'' in \emph{2024 IEEE/ACM International Conference On Computer Aided Design (ICCAD)}, 2024.

\bibitem{papamichael2015nautilus}
M.~K. Papamichael, P.~Milder, and J.~C. Hoe, ``Nautilus: Fast automated ip design space search using guided genetic algorithms,'' in \emph{Proceedings of the 52nd Annual Design Automation Conference}, 2015, pp. 1--6.

\bibitem{pan2025survey}
J.~Pan, G.~Zhou, C.-C. Chang, I.~Jacobson, J.~Hu, and Y.~Chen, ``A survey of research in large language models for electronic design automation,'' \emph{ACM Transactions on Design Automation of Electronic Systems}, 2025.

\bibitem{yao2024rtlrewriter}
X.~Yao, Y.~Wang, X.~Li, Y.~Lian, R.~Chen, L.~Chen, M.~Yuan, H.~Xu, and B.~Yu, ``Rtlrewriter: Methodologies for large models aided rtl code optimization,'' in \emph{Proceedings of the 43rd IEEE/ACM International Conference on Computer-Aided Design}, 2024, pp. 1--7.

\bibitem{chen2024dawn}
L.~Chen, Y.~Chen, Z.~Chu, W.~Fang, T.-Y. Ho, R.~Huang, Y.~Huang, S.~Khan, M.~Li, X.~Li \emph{et~al.}, ``The dawn of ai-native eda: Opportunities and challenges of large circuit models,'' \emph{arXiv preprint arXiv:2403.07257}, 2024.

\bibitem{vaswani2017attention}
A.~Vaswani, N.~Shazeer, N.~Parmar, J.~Uszkoreit, L.~Jones, A.~N. Gomez, L.~Kaiser, and I.~Polosukhin, ``Attention is all you need,'' in \emph{Proceedings of the 31st International Conference on Neural Information Processing Systems}, ser. NIPS'17.\hskip 1em plus 0.5em minus 0.4em\relax Red Hook, NY, USA: Curran Associates Inc., 2017, p. 6000–6010.

\bibitem{brown2020gpt3}
\BIBentryALTinterwordspacing
T.~Brown, B.~Mann, N.~Ryder, M.~Subbiah, J.~D. Kaplan, P.~Dhariwal, A.~Neelakantan, P.~Shyam, G.~Sastry, A.~Askell, S.~Agarwal, A.~Herbert-Voss, G.~Krueger, T.~Henighan, R.~Child, A.~Ramesh, D.~Ziegler, J.~Wu, C.~Winter, C.~Hesse, M.~Chen, E.~Sigler, M.~Litwin, S.~Gray, B.~Chess, J.~Clark, C.~Berner, S.~McCandlish, A.~Radford, I.~Sutskever, and D.~Amodei, ``Language models are few-shot learners,'' in \emph{Advances in Neural Information Processing Systems}, H.~Larochelle, M.~Ranzato, R.~Hadsell, M.~Balcan, and H.~Lin, Eds., vol.~33.\hskip 1em plus 0.5em minus 0.4em\relax Curran Associates, Inc., 2020, pp. 1877--1901. [Online]. Available: \url{https://proceedings.neurips.cc/paper_files/paper/2020/file/1457c0d6bfcb4967418bfb8ac142f64a-Paper.pdf}
\BIBentrySTDinterwordspacing

\bibitem{grattafiori2024llama}
A.~Grattafiori, A.~Dubey, A.~Jauhri, A.~Pandey, A.~Kadian, A.~Al-Dahle, A.~Letman, A.~Mathur, A.~Schelten, A.~Vaughan \emph{et~al.}, ``The llama 3 herd of models,'' \emph{arXiv preprint arXiv:2407.21783}, 2024.

\bibitem{wei2022cot}
J.~Wei, X.~Wang, D.~Schuurmans, M.~Bosma, B.~Ichter, F.~Xia, E.~H. Chi, Q.~V. Le, and D.~Zhou, ``Chain-of-thought prompting elicits reasoning in large language models,'' in \emph{Proceedings of the 36th International Conference on Neural Information Processing Systems}, ser. NIPS '22, 2022.

\bibitem{li2023structuredchainofthoughtpromptingcode}
\BIBentryALTinterwordspacing
J.~Li, G.~Li, Y.~Li, and Z.~Jin, ``Structured chain-of-thought prompting for code generation,'' 2023. [Online]. Available: \url{https://arxiv.org/abs/2305.06599}
\BIBentrySTDinterwordspacing

\bibitem{openai2024o1}
\BIBentryALTinterwordspacing
OpenAI, ``Learning to reason with llms,'' 2024. [Online]. Available: \url{https://openai.com/index/learning-to-reason-with-llms/}
\BIBentrySTDinterwordspacing

\bibitem{lewis2020rag}
P.~Lewis, E.~Perez, A.~Piktus, F.~Petroni, V.~Karpukhin, N.~Goyal, H.~K\"{u}ttler, M.~Lewis, W.-t. Yih, T.~Rockt\"{a}schel, S.~Riedel, and D.~Kiela, ``Retrieval-augmented generation for knowledge-intensive nlp tasks,'' in \emph{Proceedings of the 34th International Conference on Neural Information Processing Systems}, ser. NIPS '20, 2020.

\bibitem{petroni2020kilt}
F.~Petroni, A.~Piktus, A.~Fan, P.~S.~H. Lewis, M.~Yazdani, N.~D. Cao, J.~Thorne, Y.~Jernite, V.~Plachouras, T.~Rockt{\"{a}}schel, and S.~Riedel, ``{KILT:} a benchmark for knowledge intensive language tasks,'' \emph{CoRR}, vol. abs/2009.02252, 2020.

\bibitem{karpukhin2020dense}
V.~Karpukhin, B.~Oguz, S.~Min, L.~Wu, S.~Edunov, D.~Chen, and W.~Yih, ``Dense passage retrieval for open-domain question answering,'' \emph{CoRR}, vol. abs/2004.04906, 2020.

\bibitem{zhang2022fastbo}
\BIBentryALTinterwordspacing
Z.~Zhang, T.~Chen, J.~Huang, and M.~Zhang, ``A fast parameter tuning framework via transfer learning and multi-objective bayesian optimization,'' in \emph{Proceedings of the 59th ACM/IEEE Design Automation Conference}, ser. DAC '22.\hskip 1em plus 0.5em minus 0.4em\relax New York, NY, USA: Association for Computing Machinery, 2022, p. 133–138. [Online]. Available: \url{https://doi.org/10.1145/3489517.3530430}
\BIBentrySTDinterwordspacing

\bibitem{cui2025curie}
\BIBentryALTinterwordspacing
H.~Cui, Z.~Shamsi, G.~Cheon, X.~Ma, S.~Li, M.~Tikhanovskaya, P.~C. Norgaard, N.~Mudur, M.~B. Plomecka, P.~Raccuglia, Y.~Bahri, V.~V. Albert, P.~Srinivasan, H.~Pan, P.~Faist, B.~A. Rohr, M.~J. Statt, D.~Morris, D.~Purves, E.~Kleeman, R.~Alcantara, M.~Abraham, M.~Mohammad, E.~P. VanLee, C.~Jiang, E.~Dorfman, E.-A. Kim, M.~Brenner, S.~S. Ponda, and S.~Venugopalan, ``{CURIE}: Evaluating {LLM}s on multitask scientific long-context understanding and reasoning,'' in \emph{The Thirteenth International Conference on Learning Representations}, 2025. [Online]. Available: \url{https://openreview.net/forum?id=jw2fC6REUB}
\BIBentrySTDinterwordspacing

\bibitem{garrido2020dealing}
E.~C. Garrido-Merch{\'a}n and D.~Hern{\'a}ndez-Lobato, ``Dealing with categorical and integer-valued variables in bayesian optimization with gaussian processes,'' \emph{Neurocomputing}, vol. 380, pp. 20--35, 2020.

\bibitem{nogueira2014bo}
\BIBentryALTinterwordspacing
F.~Nogueira, ``{Bayesian Optimization}: Open source constrained global optimization tool for {Python},'' 2014--. [Online]. Available: \url{https://github.com/bayesian-optimization/BayesianOptimization}
\BIBentrySTDinterwordspacing

\bibitem{spearman1961proof}
C.~Spearman, ``The proof and measurement of association between two things.'' 1961.

\bibitem{li2023loogle}
J.~Li, M.~Wang, Z.~Zheng, and M.~Zhang, ``Loogle: Can long-context language models understand long contexts?'' \emph{arXiv preprint arXiv:2311.04939}, 2023.

\bibitem{xu2022systematic}
F.~F. Xu, U.~Alon, G.~Neubig, and V.~J. Hellendoorn, ``A systematic evaluation of large language models of code,'' in \emph{Proceedings of the 6th ACM SIGPLAN international symposium on machine programming}, 2022, pp. 1--10.

\bibitem{liu2024lost}
N.~F. Liu, K.~Lin, J.~Hewitt, A.~Paranjape, M.~Bevilacqua, F.~Petroni, and P.~Liang, ``Lost in the middle: How language models use long contexts,'' \emph{Transactions of the Association for Computational Linguistics}, vol.~12, pp. 157--173, 2024.

\bibitem{zhang2025sledselflogitsevolution}
\BIBentryALTinterwordspacing
J.~Zhang, D.-C. Juan, C.~Rashtchian, C.-S. Ferng, H.~Jiang, and Y.~Chen, ``Sled: Self logits evolution decoding for improving factuality in large language models,'' 2025. [Online]. Available: \url{https://arxiv.org/abs/2411.02433}
\BIBentrySTDinterwordspacing

\bibitem{obrien2023contrastivedecodingimprovesreasoning}
\BIBentryALTinterwordspacing
S.~O'Brien and M.~Lewis, ``Contrastive decoding improves reasoning in large language models,'' 2023. [Online]. Available: \url{https://arxiv.org/abs/2309.09117}
\BIBentrySTDinterwordspacing

\end{thebibliography}

\end{document}